\documentclass[10pt,twocolumn,letterpaper]{article}

\usepackage{3dv}
\usepackage{times}
\usepackage{epsfig}
\usepackage{graphicx}
\usepackage{amsmath}
\usepackage{amssymb}

\usepackage{bm}
\usepackage{enumitem}
\usepackage{gensymb}  
\usepackage{mathtools}  
\usepackage{multirow}
\usepackage{mwe}
\usepackage{placeins}
\usepackage{tabularx}
\usepackage{xcolor}
\usepackage{url}

\usepackage[pagebackref=true,breaklinks=true,letterpaper=true,colorlinks,bookmarks=false]{hyperref}

\DeclarePairedDelimiter\norm{\lVert}{\rVert}

\DeclareMathOperator*{\argmin}{arg\,min}
\graphicspath{ {figures/} }

\threedvfinalcopy 

\def\threedvPaperID{82} 

\ifthreedvfinal\fi
\begin{document}

\title{Learning Iterative Robust Transformation Synchronization}

\author{Zi Jian Yew \qquad Gim Hee Lee\\
Department of Computer Science, National University of Singapore\\
{\tt\small \{zijian.yew, gimhee.lee\}@comp.nus.edu.sg}
}

\maketitle
\thispagestyle{empty}

\begin{abstract}
Transformation Synchronization is the problem of recovering absolute transformations from a given set of pairwise relative motions. Despite its usefulness, the problem remains challenging due to the influences from noisy and outlier relative motions, and the difficulty to model analytically and suppress them with high fidelity. In this work, we avoid handcrafting robust loss functions, and propose to use graph neural networks (GNNs) to learn transformation synchronization.
Unlike previous works which use complicated multi-stage pipelines, we use an iterative approach where each step consists of a single weight-shared message passing layer that refines the absolute poses from the previous iteration by predicting an incremental update in the tangent space. To reduce the influence of outliers, the messages are weighted before aggregation.
Our iterative approach alleviates the need for an explicit initialization step and performs well with identity initial poses.
Although our approach is simple, we show that it performs favorably against 
existing handcrafted and learned synchronization methods through experiments on both SO(3) and SE(3) synchronization.
\end{abstract}

\section{Introduction}

The reconstruction of complete scenes has many applications in virtual reality and robotics, and requires the recovery of camera motions to align multiple images or depth scans. A two-step workflow is commonly used to recover the camera motions.
In the first step, the relative motion between any two images or point clouds is recovered. Such relative motions are typically estimated using pairwise registration techniques, which is a well-studied topic for images \cite{lowe1999sift,rublee2011orb,yi2016lift} and point clouds \cite{icp,chen-medioni,FPFH,zeng20173dmatch,yew2020RPMNet}.
The second step jointly optimizes the relative motions from the first step to recover the absolute camera poses, a problem known as Transformation Synchronization. Unfortunately, the pairwise registration process in the first step can be noisy and corrupted with outliers, leading to inaccurate or wrong input relative motions. This makes Transformation Synchronization extremely challenging.

\begin{figure}[t]
\centering
\includegraphics[width=\linewidth]{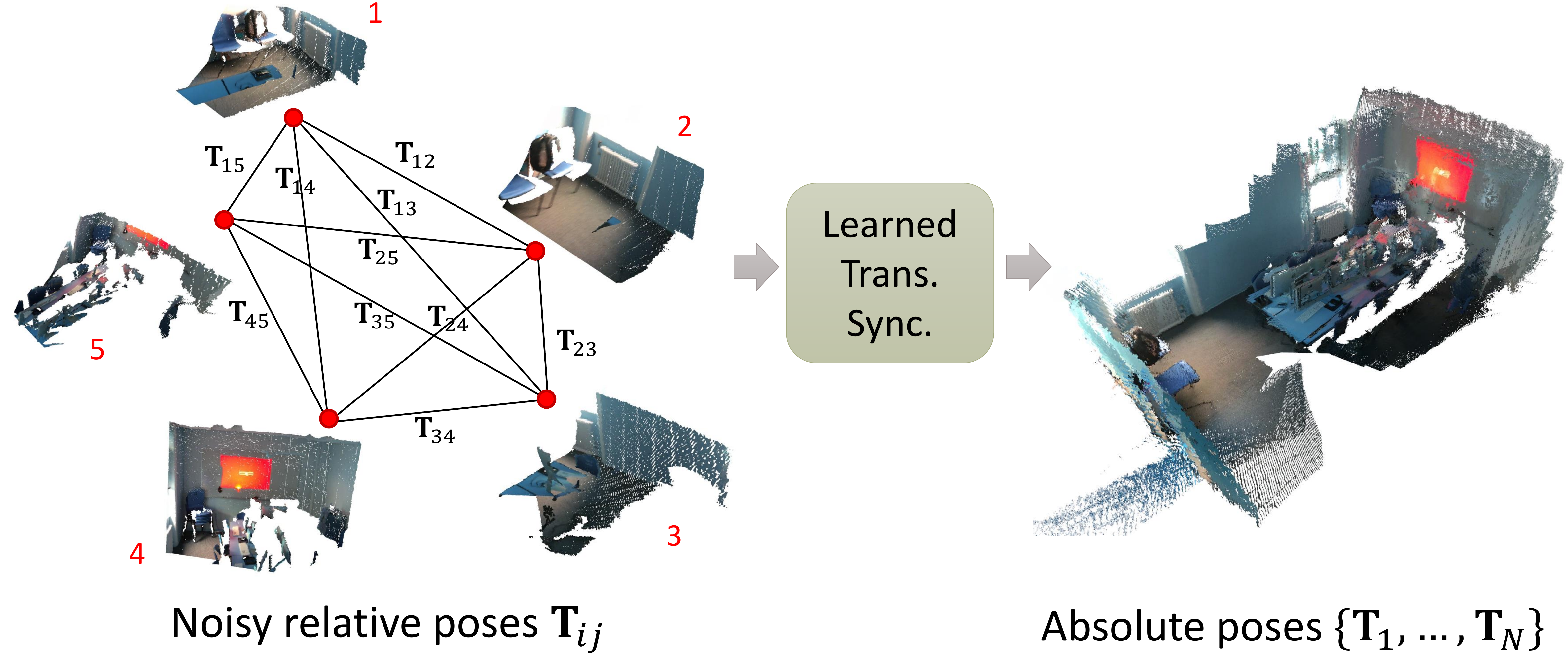}
\caption{Our learned transformation synchronization approach takes as input noisy relative poses with outliers (\eg from point cloud pairwise registration) and recovers the absolute poses.}
\label{fig:teaser}
\vspace{-2mm}
\end{figure}

Our work addresses the second problem: it takes as input a set of noisy pairwise relative motions with outliers and outputs the absolute poses (Fig. \ref{fig:teaser})
and an inlier/outlier estimate of each relative pose.
Traditional approaches to this problem \cite{chatterjee,weiszfeld} often make use of handcrafted robust loss functions, \eg Geman-McClure, to reduce the influence of noise and outliers. However, these loss functions are designed based on certain assumptions of the noise characteristics or outlier distribution of the data that may not be applicable in real-world settings. To address this, several works \cite{huang2019learn2sync,gojcic2020lmpr} incorporate learning into the problem. However, they require information from the pairwise registration (\eg point correspondences \cite{gojcic2020lmpr} or registered point clouds \cite{huang2019learn2sync}) to deduce the inlier probabilities of each relative pose. They also do not learn to optimize the poses directly, and instead feed these probabilities to a handcrafted Iterative Reweighted Least Squares (IRLS) based spectral method.

Purkait \etal \cite{purkait2019neurora} is the first work to learn to optimize the poses, using graph neural networks (GNNs) to predict the absolute poses. Due to the challenges in predicting absolute poses directly using a network, the authors used a cumbersome three-stage scheme, first cleaning the view-graph, before initializing the absolute poses over a minimal spanning tree, and finally refining the absolute poses. The use of the minimal spanning tree makes the network non end-to-end trainable, and the final stage requires good initialization from the previous stages and would otherwise give poor pose estimates. The recent work from Yang \etal \cite{yang2021endtoendAvg} proposes a differentiable multi-source propagation scheme to learn rotation averaging in an end-to-end fashion.
However, the network still requires separate stages to predict the inlier probabilities and refine the absolute poses. Similar to \cite{huang2019learn2sync,gojcic2020lmpr}, it also makes use of additional information to accurately predict the inlier probabilities of each relative pose.

In this work, we take inspiration from iterative optimization methods and propose to use a \emph{single} recurrent GNN for the entire optimization. Our network takes in a view-graph where the nodes represent the absolute poses that we wish to recover, and the edges correspond to the noisy input relative motions.
Different from \cite{purkait2019neurora}, we circumvent the difficulties in predicting absolute poses by using an iterative scheme where each iteration predicts incremental updates to the absolute poses.
The increments are parameterized in the tangent space so that the predicted poses remain within the manifold. 
To reduce the influence of outliers, we predict attention weights using a subnetwork to weigh the individual messages before aggregation.
Our resulting network does not contain multiple stages and is end-to-end trainable. It also does not require additional information from the pairwise matching and is thus compatible with any pairwise registration method or input data modality.

We demonstrate the effectiveness of our method through experiments on synthetic \cite{purkait2019neurora} and real world \cite{wilson20141dsfm} rotation averaging datasets, as well as rigid motion synchronization on a real-world multi-view point cloud registration dataset \cite{dai2017scannet}. Despite its simplicity, our network achieves competitive performance to existing works on the real world rotation averaging dataset, and state-of-the-art performance for the remaining datasets. This is despite not employing any additional tricks, \eg pre-filtering of the relative motions, or making use of additional information from pairwise matching or raw input data.
Our main contributions are:
\vspace{-1mm}
\begin{itemize}[itemsep=0.3em,parsep=0em]
    \item A single-stage iterative GNN for transformation synchronization which does not require initialization nor additional sources of information.
    \item Predict increments in tangent space to avoid the challenges in directly predicting absolute poses.
    \item Messages are weighted for robustness to outliers.
    \item We demonstrate the effectiveness of our algorithm on both rotation and rigid motion synchronization.
\end{itemize}
Our source code is available at \url{https://github.com/yewzijian/MultiReg}.

\section{Related Works}

\subsection{Traditional Transformation Synchronization}
Typical approaches to transformation synchronization include finding consistent cycles \cite{huber2003fully,huang2006reassembling,zach2010loopconstraints}, view-graph optimization \cite{choi2015robust,chatterjee,govindu2004lie}, low-rank matrix recovery \cite{arrigoni2018lowrank,arrigoni2016global}, and spectral methods \cite{arrigoni2016eigse3,bernard2015transsync}.
In many works, robustness to outliers is achieved through robust loss functions. For example, \cite{weiszfeld} uses a $\ell_1$-norm based cost to average relative rotations using the Weiszfeld algorithm. Chatterjee and Govindu \cite{chatterjee} first use $\ell_{1}$ loss before switching to a more robust $\ell_{1/2}$ or Geman-McClure loss function in an iteratively reweighted least squares (IRLS) manner. Some approaches also require a good initialization to avoid local minima, \eg by making use of odometry constraints between consecutive frames which tend to be outlier-free \cite{choi2015robust,lan2019robust}.
In this work, we avoid defining a specific loss function but instead use a data-driven approach to learn to optimize robustly. We also do not assume any specific initialization scheme and can perform well with identity initialization.

\subsection{Learned Synchronization Methods}
Several works learn to robustly synchronize transformations from data. Huang \etal \cite{huang2019learn2sync} incorporate a learned component to predict the per-iteration weights for input to an IRLS optimization scheme. Gojcic \etal \cite{gojcic2020lmpr} extend the work with a learned point cloud registration module. Both these works make use of extra information from the point clouds being registered, and still rely on a handcrafted synchronization method.
NeuRoRA \cite{purkait2019neurora} learns to synchronize rotations using GNNs via a three-stage approach: 1) clean up outliers, 2) bootstrap initial absolute rotations, and 3) refine absolute rotations. The bootstrapping in the second step is required to initialize the absolute poses near their correct values, so the network in the third step only has to refine them.
Recently, Yang \etal \cite{yang2021endtoendAvg} makes use of a differentiable propagation scheme to allow end-to-end training, but the network still requires multiple stages to clean, initialize and refine the poses. It also requires additional inputs from the images and correspondences.
Our work is similar in spirit to \cite{purkait2019neurora} and also uses GNNs for the optimization. However, we use a single recurrent GNN to incrementally refine from identity poses, avoiding the need for an explicit initialization stage. Moreover, we demonstrate our approach on both rotation and rigid motion synchronization on image and point cloud datasets.

\subsection{Graph Neural Networks}
Graph Neural Networks (GNNs) \cite{Duvenaud2015_NeuralFPs,kipf2016semi} generalizes the convolution operations of typical convolution neural networks to graphs which generally have irregular non-Euclidean structures. Message Passing Neural Networks (MPNNs) \cite{gilmer2017quantum} and Graph Networks \cite{battaglia2018relational} extends vanilla GNNs by incorporating edge and global graph information during aggregation.
Recurrent GNN models \cite{gori2005GNNs} share the same network parameters in each step. Graph Attention Networks \cite{velivckovic2017gat} incorporate attention mechanisms to allow each node to focus on relevant neighboring nodes. 
In our work, we use a recurrent scheme, where each iteration represents a pose refinement step. We also incorporate an attention mechanism to weigh incoming edges to reject the influence from outlier edges.

\begin{figure*}[t]
\centering
\includegraphics[width=\linewidth]{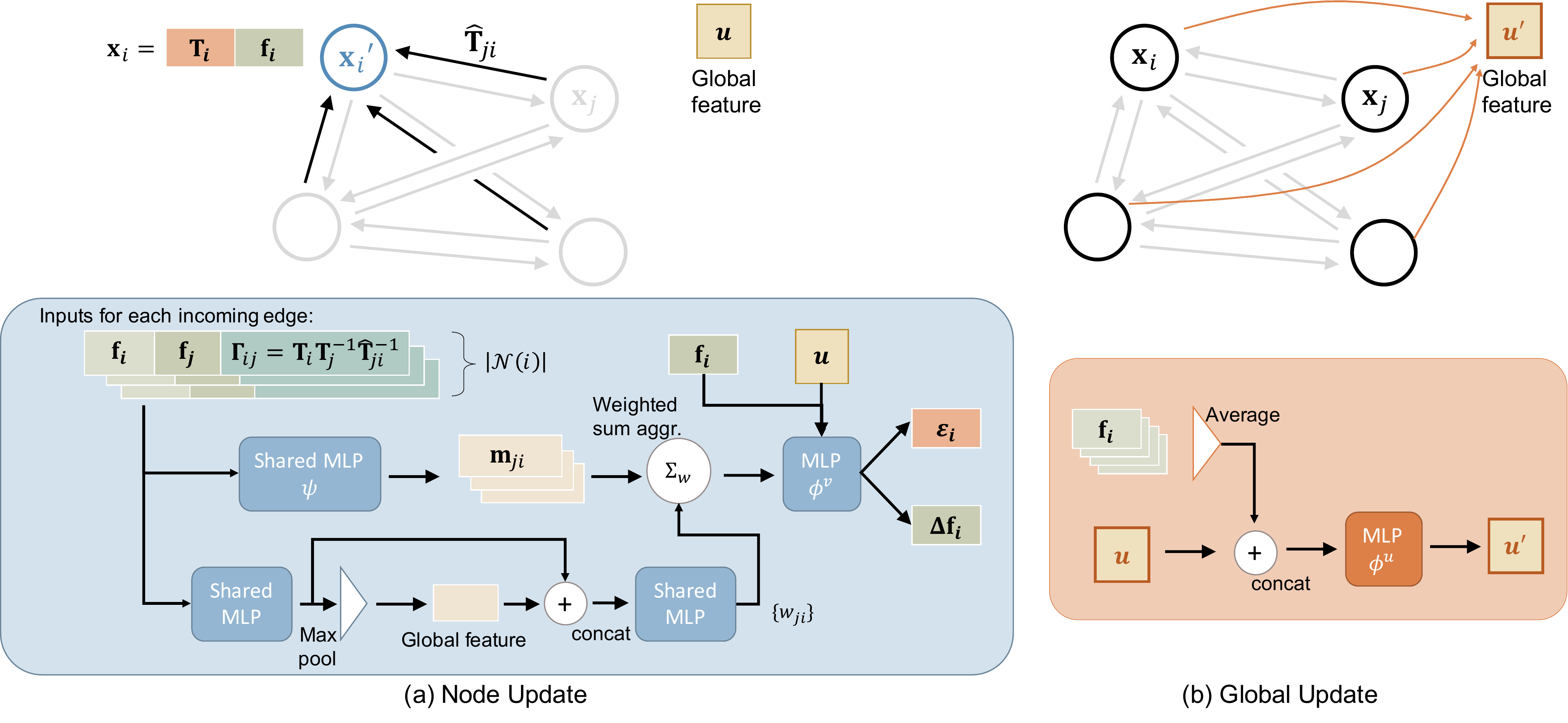}
\caption{Our iterative network architecture. Each iteration performs message passing to update (a) node and (b) global graph features.}
\label{fig:network-arch}
\end{figure*}

\section{Problem Formulation}
The objective of transformation synchronization is to recover the absolute camera poses $\mathbf{T}_i, \forall i \in \{1, ..., N\}$ from relative pairwise motions $\hat{\mathbf{T}}_{ij}$.
For 3D rotation synchronization, the absolute and relative motions are elements of the Special Orthogonal Group $\mathbf{SO(3)}$. For rigid motions the Special Euclidean Group $\mathbf{SE(3)}$ group is used, which comprises a rotation $\mathbf{R}_i \in \mathbf{SO(3)}$ and translation $\mathbf{t}_i \in \mathbb{R}^3$.

We can express the information as a view-graph $\mathcal{G}=\{\mathcal{V}, \mathcal{E}\}$, where each vertex $v \in \mathcal{V}$ corresponds to an absolute pose, and each edge $(i, j) \in \mathcal{E}$ represents the relative transformation $\hat{\mathbf{T}}_{ij}$ between cameras $i$ and $j$ associated with it. The relative and absolute transforms in the view-graph are related by the constraint:
\begin{equation} \label{eq:consistency}
    \hat{\mathbf{T}}_{ij} = \mathbf{T}_{i} \mathbf{T}_{j}^{-1}, \quad \forall(i,j) \in \mathcal{E}.
\end{equation}

In practice, these relative transformations are typically obtained from pairwise registration algorithms, \eg \cite{lowe1999sift,icp,FPFH}, and are corrupted by noise and outliers. As a result, a solution $\mathbf{T}_{\mathcal{V}} = \{\mathbf{T}_1, ..., \mathbf{T}_N\}$ that exactly satisfies all constraints in Eq. (\ref{eq:consistency}) cannot be found. To circumvent this problem, transformation synchronization seeks to recover the absolute transformations with minimum consistency error:
\begin{equation} \label{eq:opt-d}
    \argmin_{\{\mathbf{T}_1, \dots \mathbf{T}_N\}}  \sum_{(i,j) \in \mathcal{E}} \rho \left(d (\hat{\mathbf{T}}_{ji}, \mathbf{T}_{i} \mathbf{T}_{j}^{-1}) \right),
\end{equation}
where $d(\cdot, \cdot)$ is a distance measure between the two transformations. $\rho(\cdot)$ is a robust loss function to minimize the influence of outliers.

\section{Our Approach}
Instead of handcrafting robust loss functions and distance measures, we use a GNN to learn to perform transformation synchronization, as illustrated in Fig. \ref{fig:network-arch}.
GNNs performs message passing from one node to another along edges to propagate information throughout a directed graph. Let $\mathbf{x}_i^{k}$ denote the node features of node $i$ after the $k^{\text{th}}$ message passing, then in its general form, the nodes features are updated using message passing as follows:
\begin{equation}\label{eq:node-update}
  \mathbf{x}_i^{k+1} \leftarrow \phi^v
    \left(
    \mathbf{x}_i^{k},
    \mathbf{u}^{k},
    \square_{j\in\mathcal{N}(i)} \mathbf{m}_{ji}^{k}
    \right),
\end{equation}
where $\mathbf{u}^k$ is a global graph feature providing global context, which we compute by aggregating over all node features:
\begin{equation}\label{eq:global-update}
  \mathbf{u}^{k+1} \leftarrow \phi^g
    \left(
    \mathbf{u}^{k},
    \frac{1}{\norm{\mathcal{V}}}\sum_{i \in \mathcal{V}}\mathbf{x}_i^{k+1},
    \right).
\end{equation}
$\square$ denotes a permutation invariant aggregation function (Section \ref{sect:aggr-scheme}) that aggregates the messages $\mathbf{m}_{ji}$ from all neighboring nodes $j\in\mathcal{N}(i)$. $\phi^v, \phi^g$ denote arbitrary functions, which we use multilayer perceptrons (MLPs). To avoid notation clutter, we omit the iteration indices $k$ in the subsequent sections. The messages $\mathbf{m}_{ji}$ from node $j$ to $i$  are computed by considering the edge features $\mathbf{e}_{ji}$ and the features of adjacent nodes:
\begin{equation}\label{eq:message1}
    \mathbf{m}_{ji} = \psi(\mathbf{x}_i, \mathbf{x}_j, \mathbf{e}_{ji}),
\end{equation}
where $\psi(\cdot)$ is another MLP. 

In our work, we use an iterative approach where each iteration consists of a single message passing layer that refines the absolute poses from the previous step. The iterations are recurrent in nature: each iteration uses the same message passing layer with shared weights. This differs from prior works \cite{purkait2019neurora,yang2021endtoendAvg} which seek to output the absolute poses in a single pass through a GNN consisting of several non-weight shared message passing layers, and as a result require a separate initialization mechanism. We now describe our message passing operation in further detail.

\subsection{Predicting Increments in Tangent Space}
It can be challenging to predict absolute poses since the values can span a large range, \eg translation values can be as large as the size of the scene. We circumvent this by predicting increments on the absolute poses. To ensure the predicted poses remain on the $\mathbf{SO(3)}$ or $\mathbf{SE(3)}$ manifold, the pose increments $\bm{\varepsilon}_i$ are predicted on the corresponding tangent space, \ie $\mathfrak{so(3)}$ or $\mathfrak{se(3)}$, which can be applied to update the poses as:
\begin{equation}
    \mathbf{T}_i \leftarrow \mathbf{T}_i \boxplus \bm{\varepsilon}_i,
\end{equation}
where the box-plus $\boxplus$ operator is a generalization of the addition operator $+$ in Euclidean space, and is defined for Lie Groups such as $\mathbf{SO(3)}$ and $\mathbf{SE(3)}$ as:
\begin{align}
    \mathbf{T}_i \boxplus \bm{\varepsilon}_i = \exp \left({\bm{\hat{\varepsilon}_i}} \right) \mathbf{T}_i,
\end{align}

Since we are predicting only the pose increments at each iteration, the measured relative poses are not directly useful, and we instead make use of the current iteration's pose residuals $\bm{\Gamma}_{ij} = \mathbf{T}_{i}\mathbf{T}_{j}^{-1}\hat{\mathbf{T}}_{ij}^{-1}$. To facilitate computation of the pose residuals, we explicitly maintain a copy of the absolute poses as part of the node feature. Specifically, our node feature $\mathbf{x_i}$ consists of two parts: the absolute pose $\mathbf{T}_i$ and a latent feature $\mathbf{f}_i$. The messages in Eq. \ref{eq:message1} thus become:
\begin{equation}
    \mathbf{m}_{ji} = \psi \left( \mathbf{f}_i, \mathbf{f}_j, \bm{\Gamma}_{ij} \right)
\end{equation}

In summary, each iteration consists of the following node update (Eq. \ref{eq:overall-node}) and global update operations (Eq. \ref{eq:overall-global}):
\begin{subequations}\label{eq:overall-node}
\begin{equation}\label{eq:overall-node1}
    (\bm{\varepsilon}_i, \Delta\mathbf{f}_i) \leftarrow \phi^v
    \left(
    \mathbf{f}_i,
    \mathbf{u},
    \square_{j\in\mathcal{N}(i)} \psi \left( \mathbf{f}_i, \mathbf{f}_j, \bm{\Gamma}_{ij} \right),
    \right)
\end{equation}
\begin{equation}
    \mathbf{T}_i \leftarrow \mathbf{T}_i \boxplus \bm{\varepsilon}_i, \quad
    \mathbf{f}_i \leftarrow \mathbf{f}_i + \Delta\mathbf{f}_i
\end{equation}
\end{subequations}
\begin{equation}\label{eq:overall-global}
  \mathbf{u} \leftarrow \phi^g
    \left(
    \mathbf{u},
    \frac{1}{\norm{\mathcal{V}}}\sum_{i \in \mathcal{V}}\mathbf{f}_i^{k+1},
    \right).
\end{equation}

\begin{figure}[t]
\centering
\includegraphics[width=\linewidth]{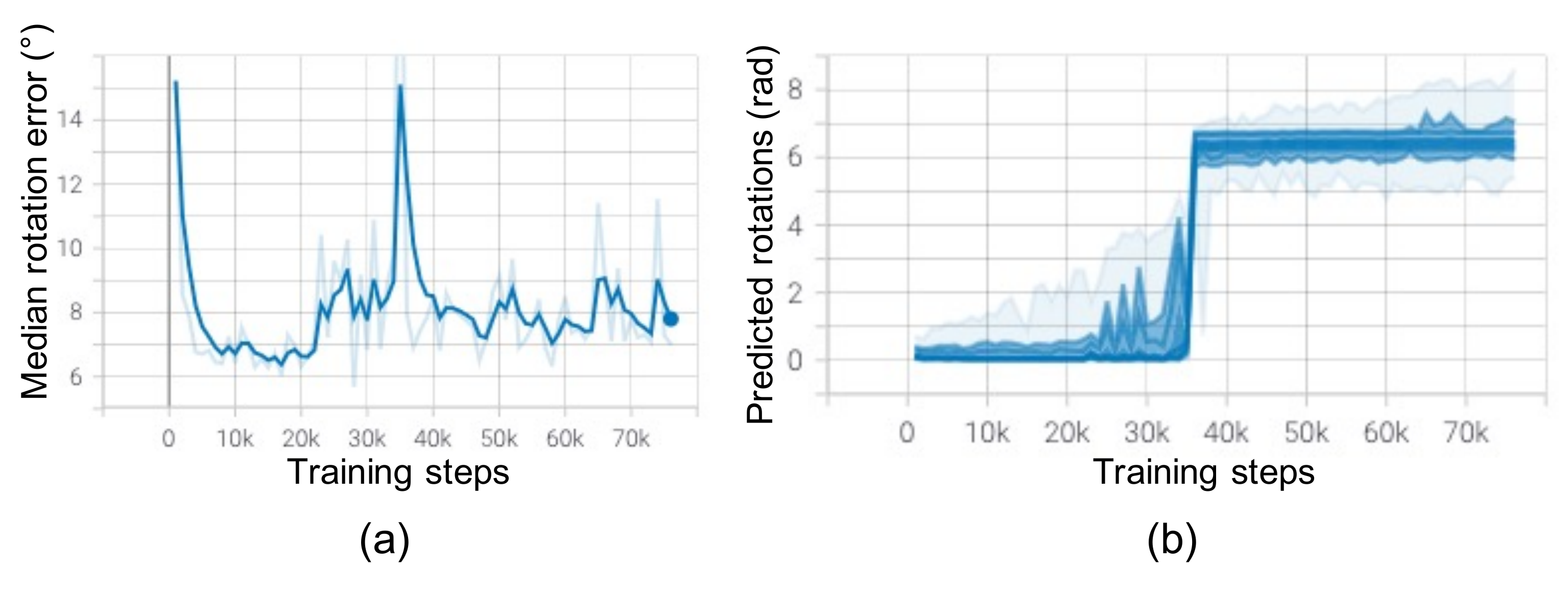}
\caption{Rotation wraparound issue. (a) Training diverged at around step 35k, as indicated by the huge increase in predicted median rotation error. (b) This is due to switching between rotation magnitudes $\theta$ and $\theta + 2\pi$.}
\label{fig:wraparound}
\end{figure}

\paragraph{Squashing rotation ranges} Empirically, we observe occasional training instability when predicting incremental updates in the tangent space (Fig. \ref{fig:wraparound}). This is due to rotation wraparound, \ie any rotations greater than $\pi$ can be represented equivalently using a rotation of at most $\pi$. We alleviate this by limiting the magnitude of the rotation update to $[0, \pi)$ by applying a squash \cite{sabour2017capsule} operation to the rotation portion of predicted increment $\bm{\varepsilon}_i$, which we denote as $\bm{\omega}_i$:
\begin{equation}\label{eq:squash}
    \bm{\omega}_i' = 
      \frac{\pi \norm{\bm{\omega_i}}^2}{1 + \norm{\bm{\omega_i}}^2} 
      \frac{\bm{\omega_i}}{\norm{\bm{\omega_i}}}.
\end{equation}
Note that here, we use the squash operation to obtain an unique rotation representation, different from \cite{sabour2017capsule} which uses it for a probabilistic interpretation.

\subsection{Weighted Aggregation Scheme}\label{sect:aggr-scheme}
The relative poses are noisy and contain outliers, and hence not all messages $\mathbf{m}_{ji}$ are useful in predicting the pose increments.
To reduce the influence from outlier relative poses, instead of a more common mean/sum aggregation, we employ a weighted sum aggregation for $\square$ in Eq. \ref{eq:overall-node1}:
\begin{equation}
    \square_{j\in\mathcal{N}(i)} \mathbf{m}_{ji}
    = \eta \left( \sum_{j \in \mathcal{N}(i)} w_{ji} \cdot \psi(\mathbf{f}_{i}, \mathbf{f}_{j}, \bm{\Gamma}_{ij}) \right),
\end{equation}
where $\eta$ denotes $\ell_2$-normalization to normalize out the effect of the possibly varying number of incoming edges.

We compute the weights $w_{ji}$ using a subnetwork that also takes in $\{\mathbf{f}_i, \mathbf{f}_j, \bm{\Gamma}_{ij}\}$ as input. Since it can be useful to consider adjacent edges when predicting where an edge corresponds to an inlier relative pose, we incorporate contextual information through a permutation invariant architecture inspired by PointNet \cite{qi2017pointnet}. Specifically, we first feed the inputs for each edge independently into a MLP, before applying max-pooling over all adjacent edges to extract a context feature. We then append the context feature with the individual edges and apply a second MLP with sigmoid activation to obtain the required weights $w_{ji}$. We also experiment with incorporating context using instance normalization \cite{moo2018learning,ulyanov2016instancenorm}, but that gave a similar performance and we retain the simpler max-pooling scheme.

\subsection{Training Loss}
Our main training loss enforces consistency over the view-graph, and is defined as the mean absolute difference between the relative motions computed from the groundtruth and predicted poses:

\begin{equation}
    \mathcal{L}_{rel}= \frac{1}{|\mathcal{E}^c|} \sum_{(i,j) \in \mathcal{E}^c}\left(
    \left| \mathbf{R}_{ij} - \mathbf{R}^{*}_{ij} \right| + 
    \lambda_t \left| \mathbf{t}_{ij} - \mathbf{t}^{*}_{ij} \right|
    \right),
\end{equation}
where $\mathbf{R}_{ij} = \mathbf{R}_{i} {\mathbf{R}_{j}}^{T}$ and $\mathbf{t}_{ij} = \mathbf{t}_i - \mathbf{R}_{ij} \mathbf{t}_{j}$ are the relative rotations and translations computed from the estimated absolute poses, and are represented as a $3 \times 3$ matrices and $3 \times 1$ vectors respectively.
$\mathbf{R}^{*}_{ij}$ and $\mathbf{t}^{*}_{ij}$ are computed similarly using the groundtruth poses.
$\lambda_{t}$ weighs the relative contributions of the rotation and translation, and we set its value to 1.
We only consider edges $(i,j) \in \mathcal{E}^c$ between nodes of the same connected component of the view-graph, which are computed by considering only edges where the angular and translation errors of the relative poses are below $15\degree$ and 0.15, respectively.
Different from \cite{purkait2019neurora}, we do not add a loss on the absolute predicted poses to enforce an unique solution: we do not find it beneficial to fix the gauge freedom as our network predicts the pose increments and not the absolute poses directly.

We also add a binary cross entropy loss $\mathcal{L}_{bce}$ on the predicted inlier weights $w_{ij}$. The groundtruth is defined as:
\begin{equation}
    w_{ij}^{*} =
    \begin{cases}
      1, \enskip \text{if } ( \text{err}_{\angle}(\hat{\mathbf{T}}_{ij}) < 5\degree ) \land 
      ( \text{err}_{t}(\hat{\mathbf{T}}_{ij}) < 0.05 ) \\
      0, \enskip \text{if } ( \text{err}_{\angle}(\hat{\mathbf{T}}_{ij}) > 15\degree ) \lor 
      ( \text{err}_{t}(\hat{\mathbf{T}}_{ij}) > 0.15 ) \\
    \end{cases},
\end{equation}
where $\text{err}_{\angle}(\cdot)$ and $\text{err}_{t}(\cdot)$ computes the rotation and translation error.
We exclude the edges that are neither labeled 0 nor 1 from the computation of $\mathcal{L}_{bce}$.

The total loss is the weighted combination of the two losses:
\begin{equation}
    \mathcal{L} = \mathcal{L}_{bce} + \lambda_{rel} \mathcal{L}_{rel},
\end{equation}
where we use a value of $\lambda_{rel}=0.2$ for all experiments. We apply the loss on the outputs in all iterations, but weigh the losses for each iteration $k$ by $\frac{1}{2}^{(K-k)}$ to give later iterations higher weights, with $K$ being the total number of optimization iterations.

\subsection{Implementation Details}
We implement our network in PyTorch and make use of the PyTorch Geometric library for the implementation of our GNN. Poses are represented as $3\times3$ and $3\times4$ matrices for rotations and rigid motions respectively. 
We initialize the global feature $\mathbf{u}$ and all node latent features $\mathbf{f_i}$ for the first iteration to all zeros, and all absolute poses $\mathbf{T}_i, \forall i \in {1, ..., N}$ to the identity transformation since we assume no prior knowledge on the absolute poses. Unless otherwise indicated, we run the message passing for $K=10$ iterations and use the pose in the final iteration for evaluation.

For all experiments, we use a dimension of 16 and 4 for $\mathbf{f}_i$ and $\mathbf{u}$, respectively. All MLPs $\phi^v, \phi^u, \psi$ are configured to have one hidden layer, concatenating the inputs where necessary, and we list their detailed configuration in the supplementary. 
The network is trained using RMSProp with a learning rate of 0.0003 with gradient clipping.

\section{Experiments}

\subsection{Synthetic SO(3) dataset}
We first evaluate on a synthetic rotation averaging dataset \cite{purkait2019neurora}, which we generate using the source code provided by the authors\footnote{\url{https://github.com/pulak09/NeuRoRA}}. The dataset is designed to mimic the characteristics in real-world datasets in terms of noise and distribution of rotation angles, and contains 1200 view-graphs which are split into 960 train, 120 validation and 120 test instances. Each view-graph contains 250-1000 cameras, with 25-50\% of edges connected by pairwise poses.

We follow the evaluation procedure in \cite{purkait2019neurora}: we first compute the best matching rotation to align the predictions to the groundtruth. We then compare the angles between the recovered $\mathbf{R_i}$ and groundtruth absolute poses $\mathbf{R_i^{*}}$. We cite the results of the baseline algorithms directly from \cite{purkait2019neurora}. However, for a fair comparison, we reran the baseline algorithms on our machine (Intel Core i7-6950X with Nvidia Titan RTX) to obtain the timings. 

The results are shown in Table \ref{table:synthetic-so3}. 
Our method outperforms all handcrafted rotation averaging algorithms \cite{chatterjee,weiszfeld,arrigoni2018lowrank,wang2013exact} and also the learned NeuRoRA \cite{purkait2019neurora}. This is despite not using an explicit initialization phase (c.f. \cite{purkait2019neurora}). Our algorithm is also faster than all handcrafted algorithms as well as the CPU timing of NeuRoRA. Although NeuRoRA has a lower GPU time, it should be noted that the timing does not include the spanning tree based bootstrapping, which takes an average of 0.09s per graph using the authors' provided MATLAB code. This is not required in our method.

\begin{table}
\small
\begin{center}
\setlength\tabcolsep{4.5pt}
\begin{tabularx}{0.9\linewidth}{X | c c c c }
  \hline
  Methods & \multicolumn{2}{c}{Angular Error} & \multicolumn{2}{c}{Time taken} \\
   & mean & median & (CPU) & (GPU) \\
  \hline
  Chatterjee \cite{chatterjee} & 2.17\degree & 1.25\degree & 6.28s & -\\
  Weiszfeld \cite{weiszfeld} & 3.35\degree & 1.02\degree & 40.19s & -\\
  Arrigoni \cite{arrigoni2018lowrank} & 2.92\degree & 1.42\degree & 4.90s & -\\
  Wang \cite{wang2013exact} & 2.77\degree & 1.40\degree & 10.61s & -\\
  NeuRoRA \cite{purkait2019neurora} & 1.30\degree & 0.68\degree & 1.60s$^{*}$ & 0.02s\textsuperscript{*} \\
  \hline
  Ours & \textbf{1.03}\degree & \textbf{0.53}\degree & 1.42s & 0.06s
  \\
  \hline
\end{tabularx}
\end{center}
\caption{Performance of rotation synchronization on NeuRoRA's synthetic dataset, and average time taken. Errors of baseline algorithms are from \cite{purkait2019neurora}. Timings are re-evaluated on our test machine. \textsuperscript{*}NeuRoRA timings exclude the bootstrapping step which takes an average of 0.09s on CPU.}
\label{table:synthetic-so3}
\end{table}

\subsection{Real SO(3) dataset}
Next, we evaluate on the 1DSfM \cite{wilson20141dsfm} dataset, which contains 15 scenes with groundtruth poses obtained from incremental SfM. Due to the limited number of scenes, we follow \cite{purkait2019neurora} 
and fine-tune network parameters from the above synthetic dataset in a leave-one-out manner.
We use $K=20$ iterations for this dataset since the view-graphs are larger.

The results are shown in Table \ref{table:rot-real}, where we compare with the same baseline algorithms as the previous section. We do not include algorithms such as \cite{yang2021endtoendAvg} which require additional information from the input images. 
Our method achieves the lowest median rotation error for 8 of the 15 scenes, and outperforms handcrafted algorithms for 3 of the remaining scenes. We note that our algorithm underperforms for 4 of the scenes (Roman Forum, Tower London, Trafalgar, Union Square). The view-graphs for these scenes tend to have sparse edges ($<15\%$ edges measured). Handling sparse view-graphs is a current limitation of our algorithm, since information requires more hops to be propagate throughout the entire graph using local message passing iterations.

\begin{table*}
\small
\begin{center}
\setlength\tabcolsep{4.5pt}
\begin{tabularx}{\linewidth}{X c c | c c c c c c c c c c | c c }
  \hline
  \multicolumn{3}{c|}{Datasets} & \multicolumn{2}{c}{Chatterjee\cite{chatterjee}} & \multicolumn{2}{c}{Weiszfeld\cite{weiszfeld}} & \multicolumn{2}{c}{Arrigoni\cite{arrigoni2018lowrank}} & \multicolumn{2}{c}{Wang\cite{wang2013exact}}
  & \multicolumn{2}{c|}{NeuRoRA\cite{purkait2019neurora}} & \multicolumn{2}{c}{Ours} \\
  Name & \# Camera & \# Edges & mn & md & mn & md & mn & md & mn & md 
  & mn & md & mn & md \\
  \hline
  Alamo & 627(577) & 49.5\% 
  & \underline{4.2} & \underline{1.1} & 4.9 & 1.4 & 6.2 & 1.6 & 5.3 & 1.4
  & 4.9 & 1.2 & \textbf{3.3} & \textbf{1.0}
  \\
  Ellis Island & 247(227) & 66.8\%
  & 2.8 & 0.5 & 4.4 & 1.0 & 3.9 & 1.2 & 3.6 & 1.1
  & \underline{2.6} & \underline{0.6} & \textbf{2.4} & \textbf{0.5}
  \\
  Gendarmenmarkt & 742(677) & 17.5\%
  & 37.6 & 7.7 & 29.4 & 9.6 & 41.6 & 13.3 & 32.6 & 6.1
  & \textbf{4.5} & \textbf{2.9} & \underline{22.8} & \underline{6.0}
  \\
  Madrid Metrop. & 394(341) & 30.7\%
  & 6.9 & \underline{1.2} & 7.5 & 2.7 & 6.0 & 1.7 & 5.0 & 1.4
  & \textbf{2.5} & \textbf{1.1} & \underline{4.3} & \underline{1.2}
  \\
  Montreal Notre. & 474(450) & 46.8\%
  & 1.5 & 0.5 & 2.1 & 0.7 & 4.8 & 0.9 & 2.0 & 0.8
  & \textbf{1.2} & \underline{0.6} & \textbf{1.2} & \textbf{0.5}
  \\
  NYC Library & 376(332) & 29.3\%
  & 3.0 & 1.3 & 3.8 & 2.1 & 3.9 & 1.5 & 2.9 & 1.4
  & \textbf{1.9} & \underline{1.1} & \underline{2.4} & \textbf{1.0}
  \\
  Notre Dame & 553(553) & 68.1\%
  & 3.5 & \textbf{0.6} & 4.7 & 0.8 & 3.9 & 1.0 & 3.5 & 0.9
  & \textbf{1.6} & \textbf{0.6} & \underline{3.1} & \textbf{0.6}
  \\
  Piazza del P. & 354(338) & 39.5\%
  & 4.0 & 0.8 & 4.8 & 1.3 & 10.8 & 1.2 & 6.2 & 1.1
  & \underline{3.0} & \textbf{0.7} & \textbf{2.5} & \textbf{0.7}
  \\
  Piccadilly & 2508(2152) & 10.2\%
  & 6.9 & 2.9 & 26.4 & 7.5 & 22.0 & 9.7 & 10.1 & 3.9
  & \textbf{4.7} & \underline{1.9} & \underline{4.8} & \textbf{1.4}
  \\
  Roman Forum & 1134(1084) & 10.9\%
  & \underline{3.1} & \underline{1.5} & 4.5 & 1.8 & 13.2 & 8.2 & 4.6 & 3.5
  & \textbf{2.3} & \textbf{1.3} & 23.8 & 2.7
  \\
  Tower of London & 508(472) & 18.5\%
  & 3.9 & 2.4 & 4.7 & 2.9 & 4.6 & 1.8 & \underline{2.9} & \underline{1.5}
  & \textbf{2.6} & \textbf{1.4} & 3.2 & 1.6
  \\
  Trafalgar & 5433(5058) & 4.6\%
  & \textbf{3.5} & \textbf{2.0} & 15.6 & 11.3 & 48.6 & 13.2 & 17.2 & 16.0
  & \underline{5.3} & \underline{2.2} & 65.1 & 72.0
  \\
  Union Square & 930(789) & 5.9\%
  & 9.3 & \underline{3.9} & 40.9 & 10.3 & \underline{9.2} & 4.4 & 6.8 & 3.2
  & \textbf{5.9} & \textbf{2.0} & 9.9 & 4.9
  \\
  Vienna Cathedral & 918(836) & 24.6\%
  & \underline{8.2} & \underline{1.2} & 11.7 & 1.9 & 19.3 & 2.39 & 10.1 & 1.8
  & \textbf{3.9} & 1.5 & 19.1 & \textbf{1.1}
  \\
  Yorkminister & 458(437) & 26.5\%
  & 3.5 & 1.6 & 5.7 & 2.0 & 4.5 & 1.6 & 3.5 & 1.3
  & \textbf{2.5} & \textbf{0.9} & \underline{2.8} & \underline{1.1}
  \\
  \hline
\end{tabularx}
\end{center}
\caption{Rotation averaging results on 1DSfM dataset (mn: mean, md: median). Bold and underline denote best and second best respectively.}
\label{table:rot-real}
\end{table*}

\subsection{SE(3) Synchronization on ScanNet}\label{sect:scannet-expt}
We next evaluate $\mathbf{SE(3)}$ synchronization on the ScanNet \cite{dai2017scannet} dataset, using the same 69 training scenes and 32 evaluation scenes as  \cite{huang2019learn2sync}, and an additional 30 scenes for validation. Since the number of training graphs is small, we perform training data augmentation for this dataset by jittering the absolute poses with the same transformation applied to the corresponding relative poses. We also corrupt a small proportion $(p=0.2)$ of the relative poses by replacing it with a random $\mathbf{SE(3)}$ transformation.

We follow LMPR \cite{gojcic2020lmpr} and extract FCGF \cite{choy2019fcgf} descriptors.
We then compute nearest neighbor matches for each point cloud pair and estimate the relative poses using the deep pairwise registration block (RegBlock) from \cite{gojcic2020lmpr}.
We use the same 30 keyframes for each test scene as \cite{gojcic2020lmpr}, which are extracted around 20 frames apart. We also extract keyframes for validation scenes in a similar manner. For training, we extract keyframes over the entire sequence, and select a random contiguous subset of 30 keyframes during each training step.
We also explore a setting similar to \cite{gojcic2020lmpr} and train using the 3DMatch \cite{zeng20173dmatch} dataset to evaluate how well our network generalizes.
The evaluation for this section follows \cite{gojcic2020lmpr} and considers the pairwise angular errors $\angle(\mathbf{R}^{*}_{ij}, \mathbf{R}_{ij})$ and translation errors $\norm{\mathbf{t}^{*}_{ij} - \mathbf{t}_{ij}}$ for  ``good'' pairs of frames, 
defined as pairs where the median point distance in the overlapping region after transformation using the measured pose $\mathbf{\hat{T}}_{ij}$ is below 0.05m.

Table \ref{table:scannet-fcgf} shows the performance metrics, and Fig. \ref{fig:teaser} and \ref{fig:qualitative} show example qualitative results. Additional qualitative results are included in the supplementary and video.
Our method outperforms all baseline algorithms. Particularly, our mean rotation and translation errors are 19.7\% and 21.4\% lower than the second best performing LMPR \cite{gojcic2020lmpr} when trained on 3DMatch, despite not making use of local feature correspondences. Also, note that we feed in all ${N \choose 2}$ relative poses into our network, \ie we do not prune out potentially bad edges unlike the baselines \cite{arrigoni2016eigse3,gojcic2020lmpr}. In the supplementary, we also include quantitative results when using pairwise registration results from Fast Global Registration \cite{zhou2016fgr}, and show that we continue to outperform existing works even when using the noisier relative poses.

\begin{table*}
\small
\begin{center}
\setlength\tabcolsep{3pt}
\begin{tabularx}{\linewidth}{l | X | c c c c c c | c c c c c c}
  \hline
  & Methods & \multicolumn{6}{c|}{Rotation error} & \multicolumn{6}{c}{Translation error (m)} \\
  & & 3\degree & 5\degree & 10\degree & 30\degree & 45\degree & Mean/Med. &
  0.05 & 0.1 & 0.25 & 0.5 & 0.75 & Mean/Med. \\
  \hline
  Inputs & FCGF\cite{choy2019fcgf}+RegBlock\cite{gojcic2020lmpr}
  & 32.6 & 37.2 & 41.0 & 46.5 & 49.4 & 65.9\degree/48.8\degree 
  & 25.1 & 34.1 & 40.0 & 43.4 & 46.8 & 1.37/0.94
  \\
  Inputs-reprod. & FCGF\cite{choy2019fcgf}+RegBlock\cite{gojcic2020lmpr}
  & 33.3 & 38.1 & 42.1 & 47.7 & 50.4 & 64.3\degree/42.3\degree
  & 25.3 & 34.4 & 40.8 & 44.2 & 47.9 & 1.43/0.89
  \\
  \hline
  \multirow{9}{6em}{After Sync} & EIGSE3 \cite{arrigoni2016eigse3}
  & 63.3 & 70.2 & 75.6 & 80.5 & 81.6 & 23.0\degree/1.7\degree
  & 42.2 & 58.5 & 69.8 & 76.9 & 79.7 & 0.45/0.06
  \\
  & LMPR \cite{gojcic2020lmpr} (3DMatch) 
  & 65.8 & 72.8 & 77.6 & 81.9 & 83.2 & 20.3\degree/1.6\degree
  & \textbf{48.4} & 67.2 & 76.5 & 79.7 & 82.0 & 0.42/0.05
  \\
  \cline{2-14}
  & Ours (3DMatch)
  & \textbf{68.9} & \textbf{76.9} & \textbf{84.0} & \textbf{87.7} & \textbf{88.4} & \textbf{16.3}\degree/\textbf{1.6\degree}
  & 47.6 & \textbf{68.1} & \textbf{81.4} & \textbf{84.5} & \textbf{86.9} & \textbf{0.33}/\textbf{0.05}
  \\
  & Ours (ScanNet)
  & 70.3 & 79.7 & 87.7 & 91.2 & 91.9 & 11.6\degree/1.6\degree
  & 51.6 & 73.0 & 84.1 & 88.3 & 89.5 & 0.28/0.05
  \\
  \cline{2-14}
  & Ours (ScanNet) w/ mean aggr
  & 68.5 & 78.3 & 85.0 & 88.6 & 89.0 & 15.0\degree/1.6\degree
  & 50.3 & 71.0 & 82.1 & 84.4 & 85.7 & 0.37/0.05
  \\
  & Ours (ScanNet) w/o inc. update
  & 67.2 & 77.4 & 87.3 & 90.5 & 91.5 & 12.0\degree/1.8\degree
  & 43.9 & 67.4 & 83.3 & 87.5 & 88.5 & 0.31/0.06
  \\
  & Ours (ScanNet) w/ sep. poses
  & 69.0 & 79.2 & 87.5 & 89.7 & 89.9 & 14.4\degree/1.7\degree
  & 47.1 & 68.5 & 82.2 & 86.9 & 88.5 & 0.32/0.05
  \\
  & Ours (ScanNet) w/o $\mathbf{f}_i$
  & 66.7 & 74.8 & 82.2 & 89.6 & 90.4 & 14.4\degree/1.7\degree
  & 46.9 & 66.3 & 79.5 & 85.8 & 87.6 & 0.33/0.05
  \\
  \hline
\end{tabularx}
\end{center}
\caption{Results of SE(3) synchronization on the ScanNet dataset. Results of baselines and inputs are obtained from \cite{gojcic2020lmpr}. The dataset indicated in parentheses denote the dataset used for training the synchronization step, all methods use FCGF descriptors that are trained on 3DMatch. Bold and indicate best performance when trained on the 3DMatch dataset (direct generalization)}
\label{table:scannet-fcgf}
\end{table*}

\begin{figure*}
    \centering
    \includegraphics[width=\linewidth]{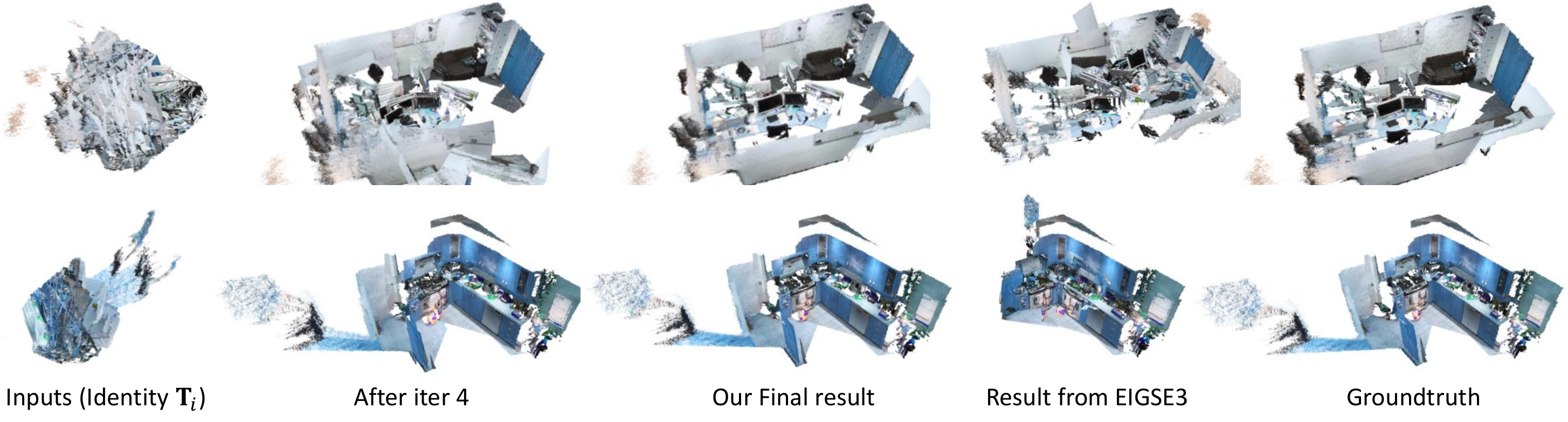}
    \caption{Qualitative rigid motion synchronization results on ScanNet (trained on ScanNet). Each view-graph consists of 30 point cloud fragments. Top: scene0025\_01, Bottom: scene0335\_02.}
    \vspace{-2mm}
    \label{fig:qualitative}
    \vspace{-1mm}
\end{figure*}

\subsection{Analysis}
We conduct further experiments in this section to understand our algorithm behavior. All experiments in this section are conducted on the ScanNet dataset.

\subsubsection{Estimation of inliers}
\begin{figure}[ht]
\centering
\includegraphics[width=\linewidth]{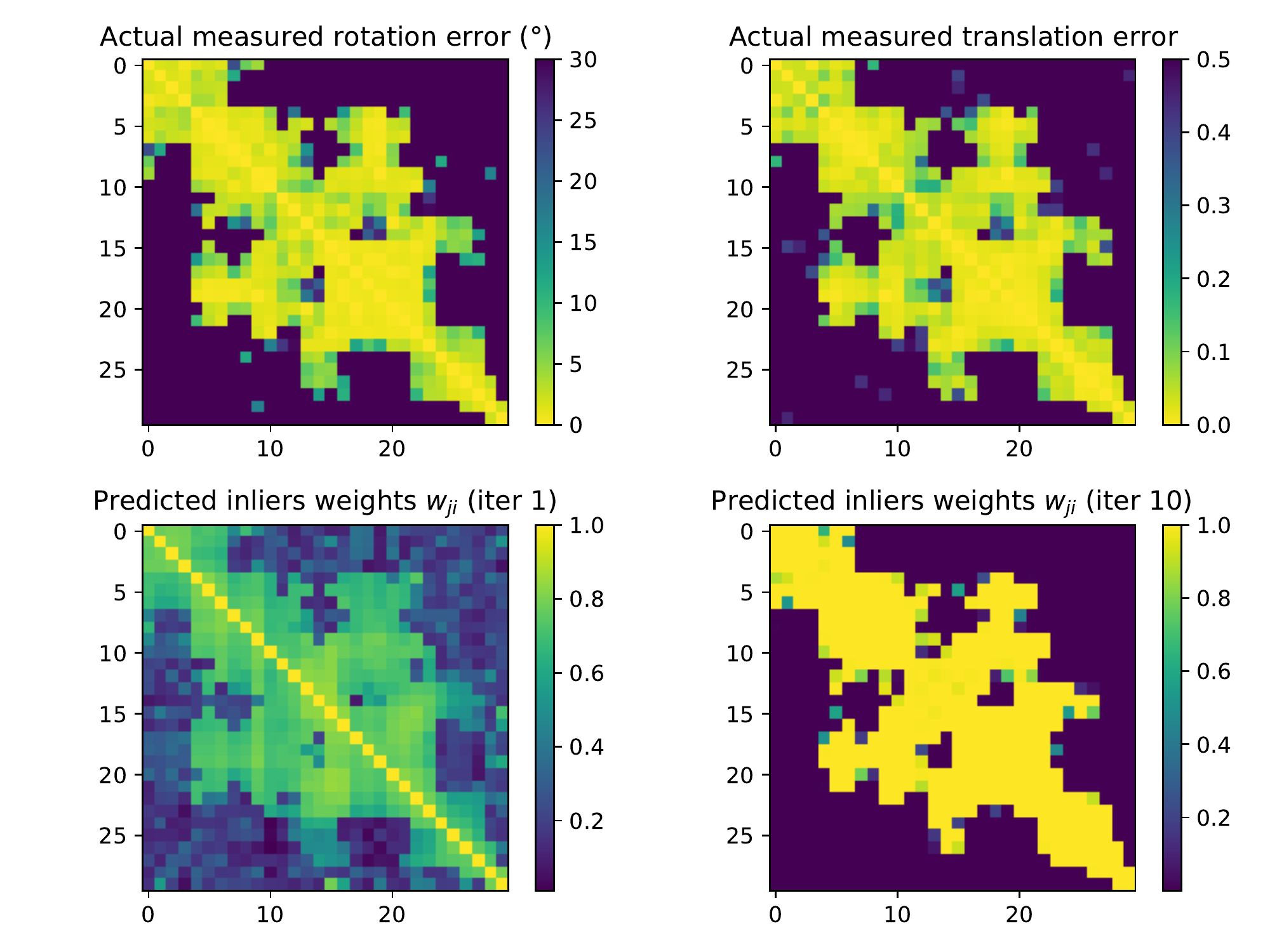}
\caption{Visualization of predicted inliers on a ScanNet scene (scene0406\_02). The top shows the measurement errors between each pair of the 30 keyframes (blue tones indicate larger errors, errors are clipped to improve contrast). The bottom row shows the predicted inlier weights $w_{ji}$ in the first and last iteration.}
\label{fig:inliers}
\end{figure}

In Fig. \ref{fig:inliers}, we show the actual angular and translation errors of the input pairwise relative motions, together with the inlier weights $w_{ji}$ predicted by our network. Interestingly, in the first iteration before any exchanges of messages, the predicted inlier weights already show some correlation to the actual measurement errors. We posit that the network has learned a prior distribution of inlier relative transformations from the data, \eg inlier relative camera motions tend to have minimal rotation in the roll axis. At the end of the optimization, the weights are well-correlated with the actual errors, which shows that the network is capable of detecting and rejecting the influence of outliers.

\begin{figure}[ht]
\centering
\includegraphics[width=\linewidth]{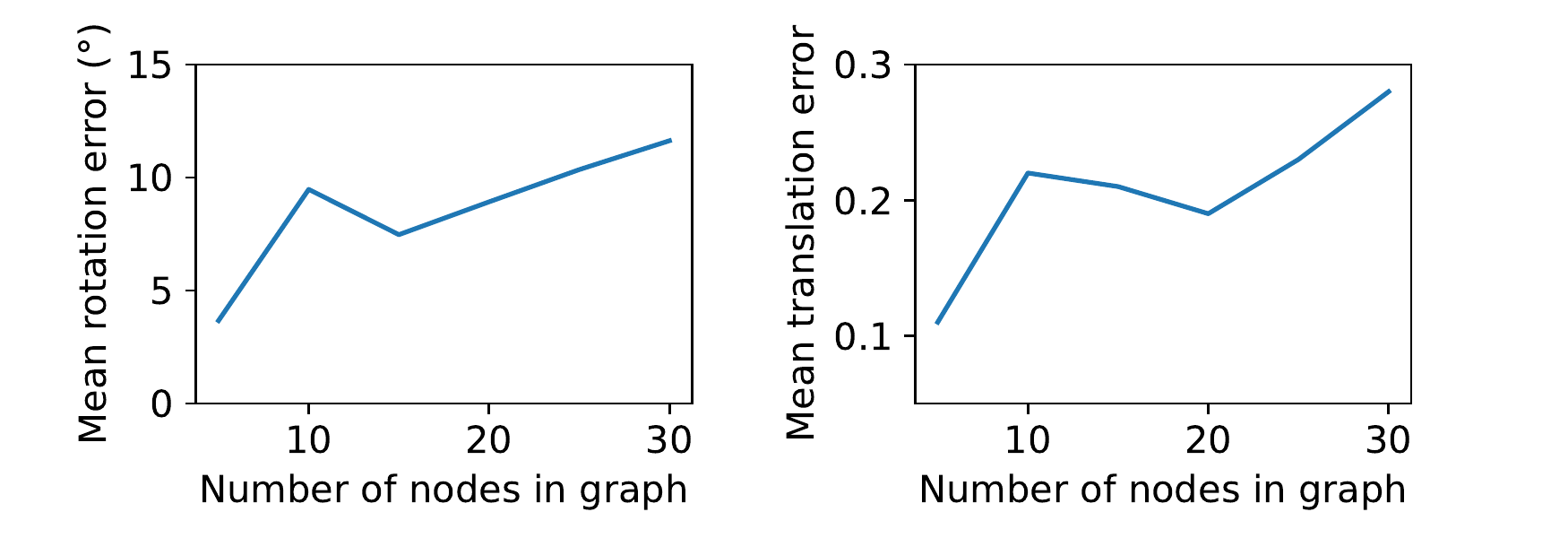}
\caption{Robustness on different sizes of view-graphs.}
\label{fig:robustness-sizes}
\end{figure}

\begin{figure}[ht]
\centering
\includegraphics[width=0.93\linewidth]{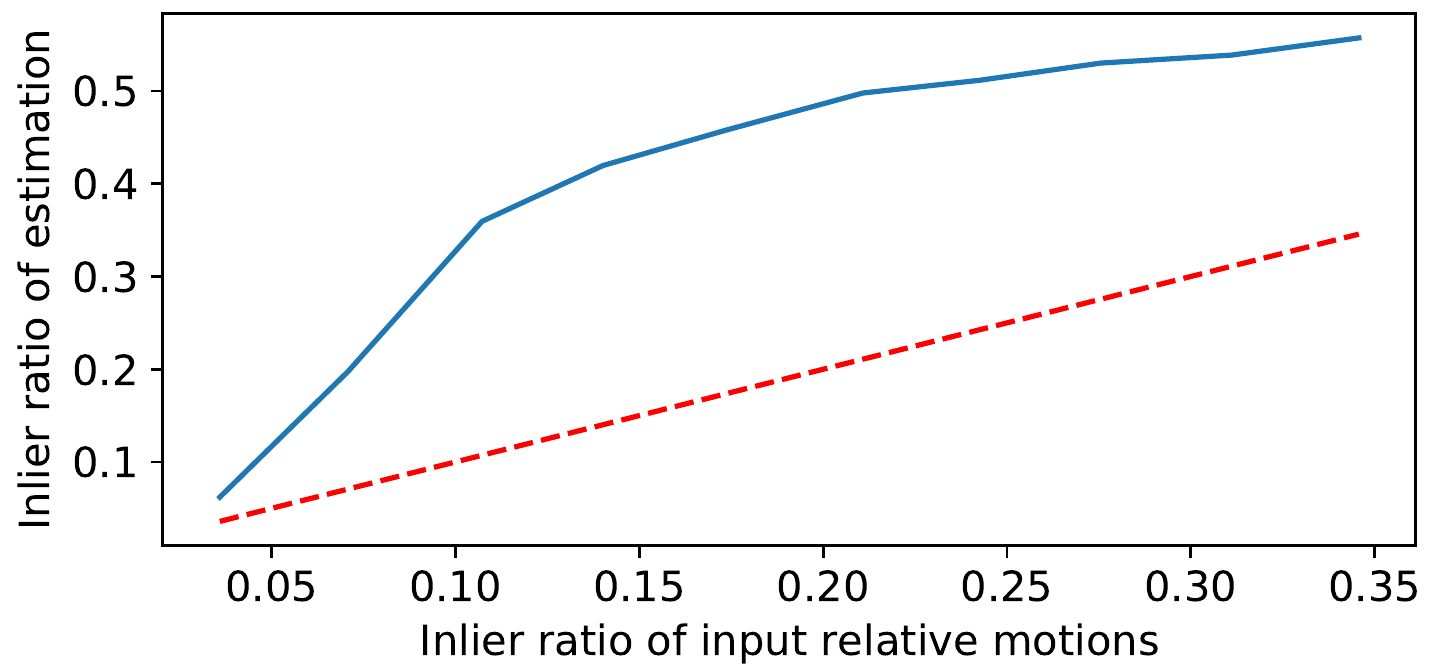}
\caption{Robustness against different inlier ratios. The red dashed line is the diagonal $y=x$. Our approach returns estimates better than the input for a wide range of inlier ratios.}
\label{fig:robustness-inliers}
\end{figure}

\subsubsection{Robustness to graphs sizes and outliers}
An important behavior for learned optimization is its generalization behavior to different view-graphs. We generate variations of the ScanNet test dataset of varying view-graph sizes by selecting a contiguous block of frames $(1, \dots , N)$ for each scene with $N$ ranging from 5 to 30, then evaluate our trained network on them. 
Fig. \ref{fig:robustness-sizes} shows that the performance remains stable for view-graphs of smaller sizes, despite the network being only trained on graphs containing exactly 30 frames.

We also analyze the robustness to outliers. We randomly corrupt varying proportions of relative motions in the ScanNet test dataset by replacing them with random $\mathbf{SE(3)}$ matrices. We then evaluate the proportion of inlier relative motion pairs for our algorithm for different input inlier ratios. Inliers are defined as motions having rotation and translation errors of below $10\degree$ and 0.1, respectively.
As seen in Fig. \ref{fig:robustness-inliers}, our algorithm improves upon the input relative poses with as little as 11\% inliers. The improvement with below 11\% inliers is less, but the predictions are still more accurate than the input noisy measurements.

\subsubsection{Number of optimization iterations}
\begin{figure}[ht]
\centering
\vspace{-3mm}
\includegraphics[width=0.97\linewidth]{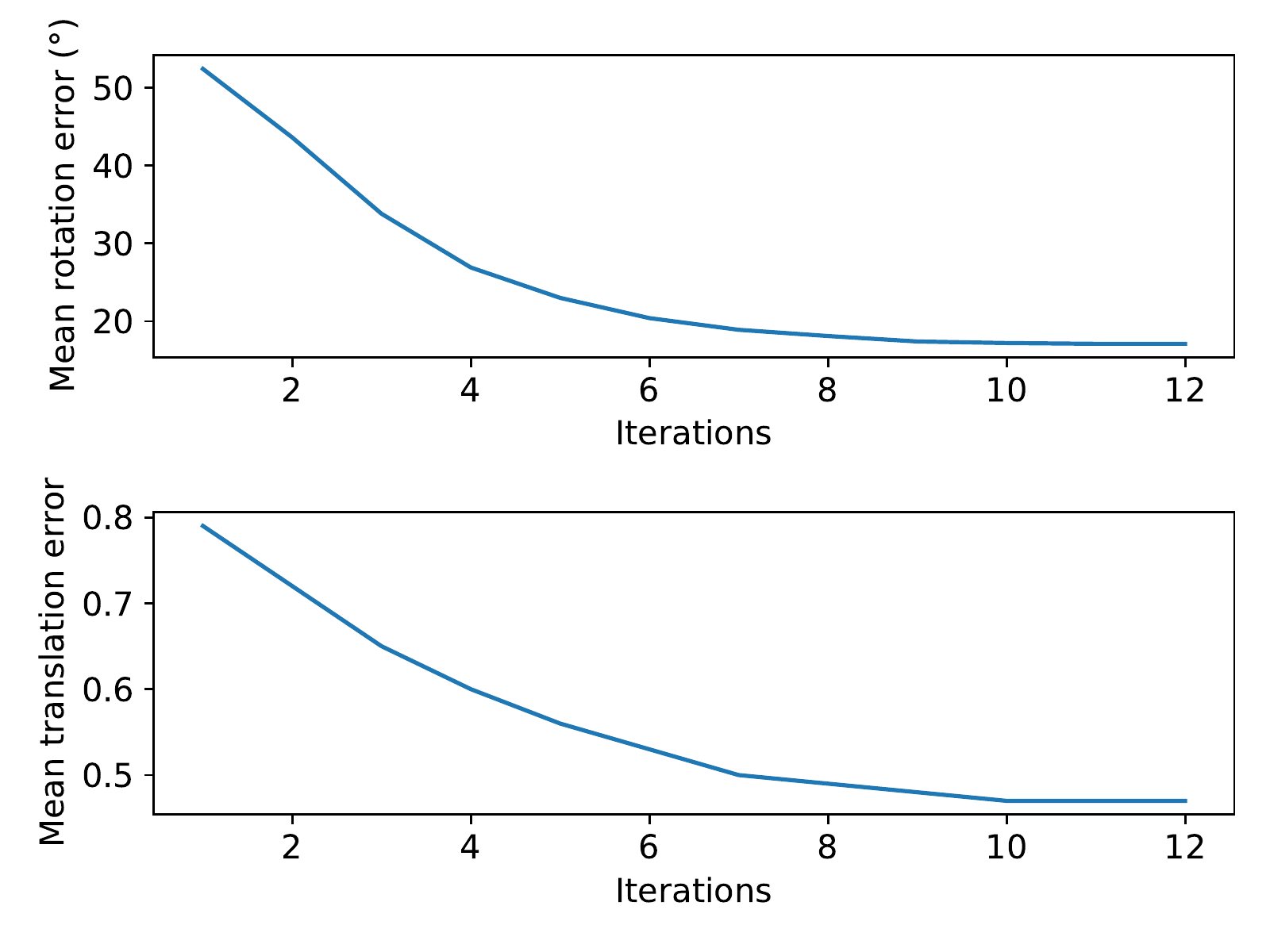}
\vspace{-3mm}
\caption{Mean rotation and translation errors for each iteration.}
\label{fig:iterations-error}
\end{figure}

Fig. \ref{fig:iterations-error} shows the mean rotation and translation errors for various iterations. We can observe that the optimization converges by around 10 iterations, which we use for most of our experiments.

\subsubsection{Ablation studies}
To understand how various choices affect the performance, we experiment with variants of our model (Rows 7-10 of Table \ref{table:scannet-fcgf}):

\begin{itemize}[itemsep=0.5em,parsep=0em]
    \item \textbf{w/ mean aggregation:} \enskip
    We replace the weighted sum aggregation with a simple mean aggregation scheme, which lead to significantly larger mean errors since the network is unable to reject the influence of outliers well.
    
    \item \textbf{w/o inc. update:} \enskip
    Instead of predicting the incremental update to the $\mathbf{SE(3)}$ poses, we directly predict the absolute pose at each iteration. Here, we use $\hat{\mathbf{T}}_{ij}$ as the edge feature since the residual motion $\bm{\Gamma}_{ij}$ is not useful for predicting the absolute poses. We see a decrease in accuracy particularly at the stricter thresholds, showing the importance of our incremental update scheme.
    
    \item \textbf{w/ sep poses:} \enskip
    We replace $\mathbf{\Gamma}_{ij}$ in the node update stage (Eq. \ref{eq:overall-node1}) with $\{\mathbf{T}_i, \mathbf{T}_j, \hat{\mathbf{T}}_{ji}\}$. Note that the resulting network is slightly bigger because of the increased number of inputs. Despite having access to the information required to reconstruct $\mathbf{\Gamma}_{ij}$, this configuration also underperformed which indicates that the residual poses $\mathbf{\Gamma}_{ij}$ are better for predicting the updates to the absolute poses.
    
    \item \textbf{w/o $\mathbf{f}_i$:} \enskip
    We exclude the latent node features $\mathbf{f}_i$ in this configuration, and the node feature consists only of the current estimated estimated pose $\mathbf{T}_i$. This leads to a decrease in accuracy over all thresholds, and indicates that useful information (on top of the current absolute poses) is stored as the node feature.
    
\end{itemize}

\section{Conclusion}
We proposed a single stage learned iterative approach to perform transformation synchronization.
Our method does not require an explicit initialization phase and works well using identity initialization. Experiments show our approach outperforms existing synchronization methods for the NeuRoRA synthetic and ScanNet datasets, and achieves competitive performance on the 1DSfM dataset. One current limitation is our reliance on local messaging passing operations causes the algorithm to underperform on large sparse graphs. Possible solutions include adding multi-hop connections \cite{yun2019graphtransformer} or coarse-to-fine approaches using graph pooling \cite{diehl2019graphpooling}, and we leave this as future work.

\vspace{-2mm}
\paragraph{Acknowledgement.}
This research is supported in part by the National Research Foundation, Singapore under its Competitive Research Program Award
NRF-CRP23-2019-0003 and the Tier 2 grant MOET2EP20120-0011 from the Singapore Ministry of Education.

{\small
\bibliographystyle{ieee_fullname}
\bibliography{references}
}

\appendix
\renewcommand\thesection{\Alph{section}}

\vspace{5mm}

\section{Exponential Maps for SO(3) and SE(3)}
We list the formulae for the exponential maps for $\mathbf{SO(3)}$ and $\mathbf{SE(3)}$ in this section. We first define the hat operator $\hat{}$ which maps a $3 \times 1$ vector to its corresponding skew-symmetric matrix:
\begin{equation}
    \bm{\omega} = 
      \begin{pmatrix}
      x \\ y \\ z
      \end{pmatrix}
    \quad\to\quad
    \hat{\bm{\omega}} = 
      \begin{pmatrix}
      0 & -z & y \\
      z & 0 & -x \\
      -y & x & 0
      \end{pmatrix}
\end{equation}

\subsection{Special Orthogonal Group SO(3)}
The exponential map for $SO(3)$ maps elements $\bm{\omega} \in \mathbb{R}^3$ to $3 \times 3$ rotation matrices. It has a closed form solution that is given by the Rodrigues' formula:

\begin{equation}\label{eq:so3exp}
    \exp(\bm{\omega}) = \mathbf{I}_3 + \frac{\sin{\theta}}{\theta}\hat{\bm{\omega}} + \frac{1 - \cos{\theta}}{\theta^2}(\hat{\bm{\omega}})^2,
\end{equation}
where $\theta = \norm{\bm{\omega}}$ is the rotation magnitude.

\subsection{Special Euclidean Group SE(3)}
Similarly, consider the $6 \times 1$ vector $\bm{\varepsilon}$:
\begin{equation}
    \bm{\varepsilon}=\begin{pmatrix}\mathbf{v} \\ \bm{\omega} \end{pmatrix} \in \mathbb{R}^6,
\end{equation}
where $\mathbf{v}\in \mathbb{R}^3 $ and $\bm{\omega} \in \mathbb{R}^3$ denote the translation and rotation components. The exponential map for $SE(3)$ maps $\bm{\varepsilon}$ to a $4 \times 4$ rigid transformation matrix, and has the following closed form:
\begin{equation}
    \exp(\bm{\varepsilon}) = 
    \begin{pmatrix}
    \exp(\bm{\omega}) & \mathbf{V v} \\
    \mathbf{0} & 1
    \end{pmatrix},
\end{equation}
where $\exp(\bm{\omega})$ is defined in Eq. \ref{eq:so3exp}, and
\begin{equation}
    \mathbf{V} = \mathbf{I}_3 + \frac{1-\cos\theta}{\theta^2} \hat{\bm{\omega}} + \frac{\theta - \sin\theta}{\theta^3}(\hat{\bm{\omega}})^2.
\end{equation}

\section{Detailed Network Architecture}
Table \ref{table:arch-se3} and \ref{table:arch-so3} shows the detailed network architecture for $\mathbf{SE(3)}$ and $\mathbf{SO(3)}$ synchronization, respectively. Note that the squash operation (Eq. \ref{eq:squash} in the main paper) is only applied to the 3 elements corresponding to the rotation component of the predicted update.

\begin{table}[t]
\small
\begin{center}
\begin{tabularx}{\linewidth}{X | c |c }
  \hline
  Stage & MLP & Architecture \\
  \hline
  Global Update & \enskip $\phi^g$ & fc(256) - ReLU - fc(4) \\
  \hline
  \multirow{2}{6em}{Node Update} & $\psi$ & fc(256) - ReLU - fc(256) - ReLU \\
  & \enskip $\phi^v$ & fc(256) - ReLU - fc(16+6) - Squash \\
  \hline
  \multirow{3}{6em}{Node Update weighting} & \multirow{3}{0.5em}{-} & fc(256) - ReLU - fc(128) \\
  & & max-pool \\
  & & fc(256)-ReLU-fc(1) \\
  \hline
\end{tabularx}
\end{center}
\caption{Detailed architecture of the MLPs in our network for SE(3) synchronization. fc($k$) denotes fully-connected layer with $k$ nodes.}
\label{table:arch-se3}
\end{table}

\begin{table}[t]
\small
\begin{center}
\begin{tabularx}{\linewidth}{X | c |c }
  \hline
  Stage & MLP & Architecture \\
  \hline
  Global Update & \enskip $\phi^g$ & fc(64) - ReLU - fc(4) \\
  \hline
  \multirow{2}{6em}{Node Update} & $\psi$ & fc(64) - ReLU - fc(64) - ReLU \\
  & \enskip $\phi^v$ & fc(64) - ReLU - fc(16+3) - Squash \\
  \hline
  \multirow{3}{6em}{Node Update weighting} & \multirow{3}{0.5em}{-} & fc(64) - ReLU - fc(32) \\
  & & max-pool \\
  & & fc(64)-ReLU-fc(1) \\
  \hline
\end{tabularx}
\end{center}
\caption{Detailed architecture of the MLPs in our network for SO(3) synchronization. fc($k$) denotes fully-connected layer with $k$ nodes.}
\label{table:arch-so3}
\end{table}

\section{Results on ScanNet using FastGR}
In this section, we show the quantitative results of $\mathbf{SE(3)}$ synchronization on ScanNet \cite{dai2017scannet} dataset using pairwise relative poses estimated using Fast Global Registration (FastGR) \cite{zhou2016fgr}. We follow the procedure in \cite{huang2019learn2sync}, and use the authors' source code\footnote{\url{https://github.com/xiangruhuang/Learning2Sync.git}} to generate the input data to our algorithm. For each scene, 100 keyframes are extracted, each 6 frames apart. FastGR is then used to estimate the pairwise relative poses between any two keyframes. 
Although we use the authors' source code, we notice that our generated data have different statistics from those reported in \cite{huang2019learn2sync} (rows 1-2 of Table \ref{table:scannet-fgr}).
We tried modifying the processing scripts to extract 30 keyframes at 20-frame interval, but the resulting statistics remain different from those reported in the paper: \eg we obtain $83.4\degree$ mean rotation error vs. the reported value of $76.3\degree$. We therefore retain the settings in the provided source code which gave mean rotational errors that are more similar.

The results are shown in Table \ref{table:scannet-fgr}. Note that the error metrics are computed on all pairwise edges for consistency with \cite{huang2019learn2sync}.
Although the relative poses from FastGR are more noisy than those estimated using the FCGF descriptors \cite{choy2019fcgf} used in the main paper, our method still substantially improves over the accuracy of the input pairwise transforms, reducing the mean rotation and translation errors by 46\% and 52\% relative to the input measurements. It also outperforms all baseline learned and handcrafted algorithms.
It should be noted that the baseline algorithms improve the rotation accuracies at the expense of translation. For example, only $2.0\%$ of relative poses predicted from L2Sync \cite{huang2019learn2sync} have translation errors below 0.05, down from $5.5\%$ in the input data. In comparison, our method improves upon the relative transforms at all thresholds for both rotation and translation.
Lastly, we should emphasize that our algorithm does not make use of additional information from the pairwise matching, unlike in \cite{huang2019learn2sync} which feeds in rendered images of the pairwise registered point clouds in order to infer whether each input relative pose is an inlier.

\begin{table*}
\small
\begin{center}
\setlength\tabcolsep{4.2pt}
\begin{tabularx}{\linewidth}{l | X | c c c c c c | c c c c c c }
  \hline
  & Methods & \multicolumn{6}{c|}{Rotation error} & \multicolumn{6}{c}{Translation error (m)} \\
  & & 3\degree & 5\degree & 10\degree & 30\degree & 45\degree & Mean/Med. &
  0.05 & 0.1 & 0.25 & 0.5 & 0.75 & Mean/Med. \\
  \hline
  Inputs & FastGR \cite{zhou2016fgr}
  & 9.9 & 16.8 & 23.5 & 31.9 & 38.4 & 76.3\degree/- & 5.5 & 13.3 & 22.0 & 29.0 & 36.3 & 1.67/-\\
  Inputs-reprod. & FastGR \cite{zhou2016fgr}
  & 20.7 & 22.2 & 24.9 & 33.0 & 37.9 & 77.4\degree/78.6\degree
  & 15.8 & 17.9 & 22.2 & 28.8 & 35.4 & 1.62/1.32\\
  \hline
  \multirow{6}{6em}{After Sync.} & Chatterjee \cite{chatterjee} + \cite{Huang2017TranslSync}
  & 6.0 & 10.4 & 17.3 & 36.1 & 46.1 & 64.4\degree/- 
  & 3.7 & 9.2 & 19.5 & 34.0 & 45.6 & 1.26/- 
  \\
  & GeoReg \cite{choi2015robust}
  & 0.2 & 0.6 & 2.8 & 16.4 & 27.1 & 87.2\degree/- 
  & 0.1 & 0.7 & 4.8 & 16.4 & 28.4 & 1.80/-
  \\
  & TranSyncV2 \cite{bernard2015transsync}
  & 0.4 & 1.5 & 6.1 & 29.0 & 42.2 & 68.1\degree/- 
  & 0.2 & 1.5 & 11.3 & 32.0 & 46.3 & 1.44/-
  \\
  & EIGSE3 \cite{arrigoni2016eigse3}
  & 1.5 & 4.3 & 12.1 & 34.5 & 47.7 & 68.1\degree/- 
  & 1.2 & 4.1 & 14.7 & 32.6 & 46.0 & 1.29/-
  \\
  & L2Sync \cite{huang2019learn2sync}
  & 34.4 & 41.1 & 49.0 & 58.9 & 62.3 & 42.9\degree/- 
  & 2.0 & 7.3 & 22.3 & 36.9 & 48.1 & 1.16/-
  \\
  \cline{2-14}
  & Ours
  & \textbf{34.7} & \textbf{41.4} & \textbf{50.7} & \textbf{62.6} & \textbf{66.5} & \textbf{41.8\degree/9.37\degree}
  & \textbf{21.6} & \textbf{31.7} & \textbf{43.1} & \textbf{54.2} & \textbf{64.0} & \textbf{0.78/0.40}
  \\
  \hline
\end{tabularx}
\end{center}
\caption{Results of SE(3) synchronization on the ScanNet dataset using pairwise poses from FastGR \cite{zhou2016fgr}. Results of baselines and inputs are obtained from L2Sync \cite{huang2019learn2sync}.  Angular and translation errors are computed by comparing pairwise poses with the groundtruth for all pairs, and we report the percentage of pairs with errors under varying thresholds as well as mean/median errors. Both L2Sync and our algorithm are trained on the ScanNet dataset, the rest are non-learning algorithms. Note that our results are based on our data (row 2) generated from the source codes of \cite{huang2019learn2sync} and have different input statistics from that reported in \cite{huang2019learn2sync} (row 1). See text for details.}
\label{table:scannet-fgr}
\end{table*}

\section{Additional Qualitative Results for ScanNet}
\begin{figure*}
    \centering
    \includegraphics[width=\linewidth]{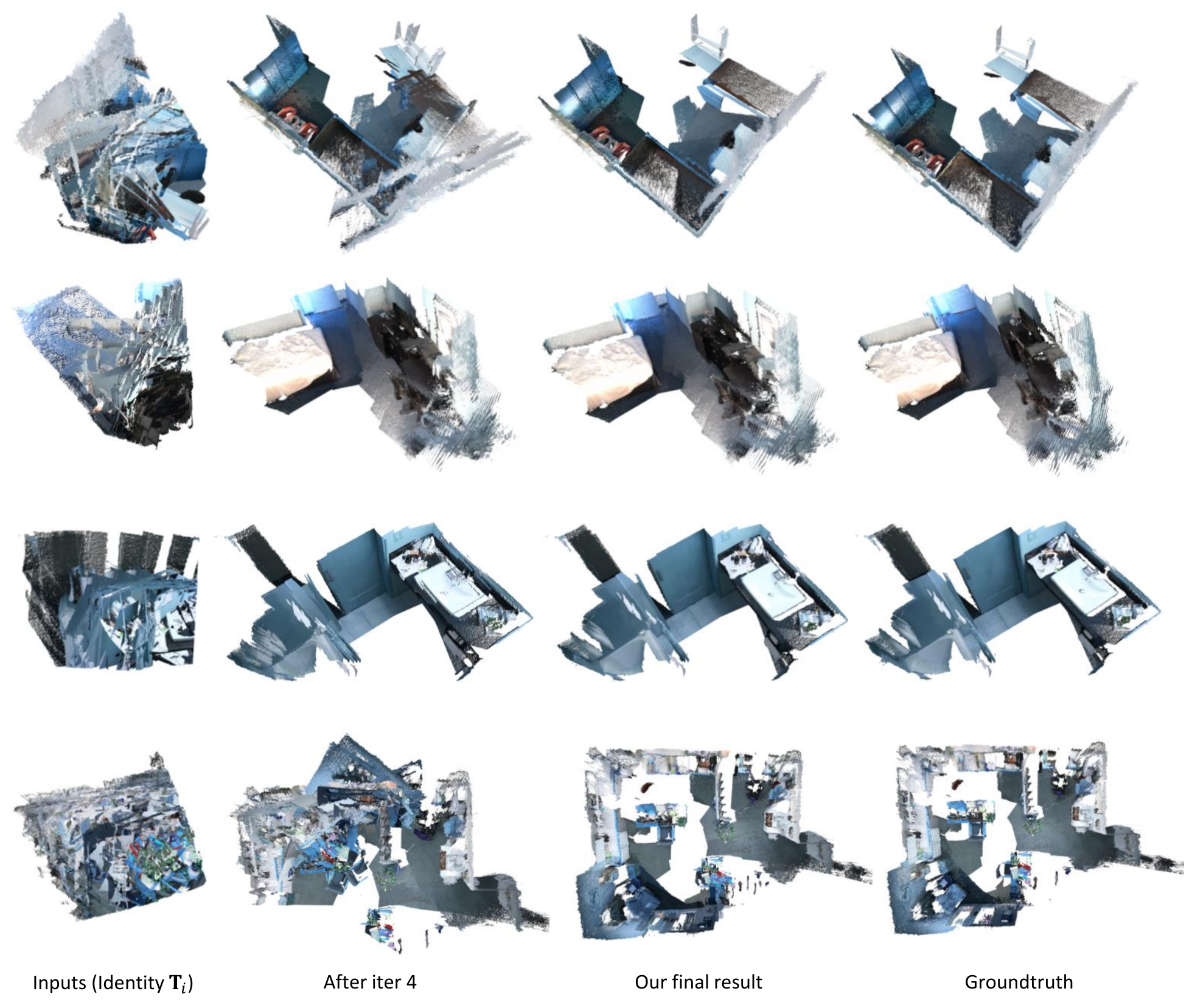}
    \caption{Additional successful cases by our algorithm using FCGF descriptors, as described in main paper. From top to bottom: scene0223\_00, scene0642\_02, scene0406\_02, scene0309\_00.}
    \label{fig:success}
\end{figure*}

\begin{figure*}
    \centering
    \includegraphics[width=\linewidth]{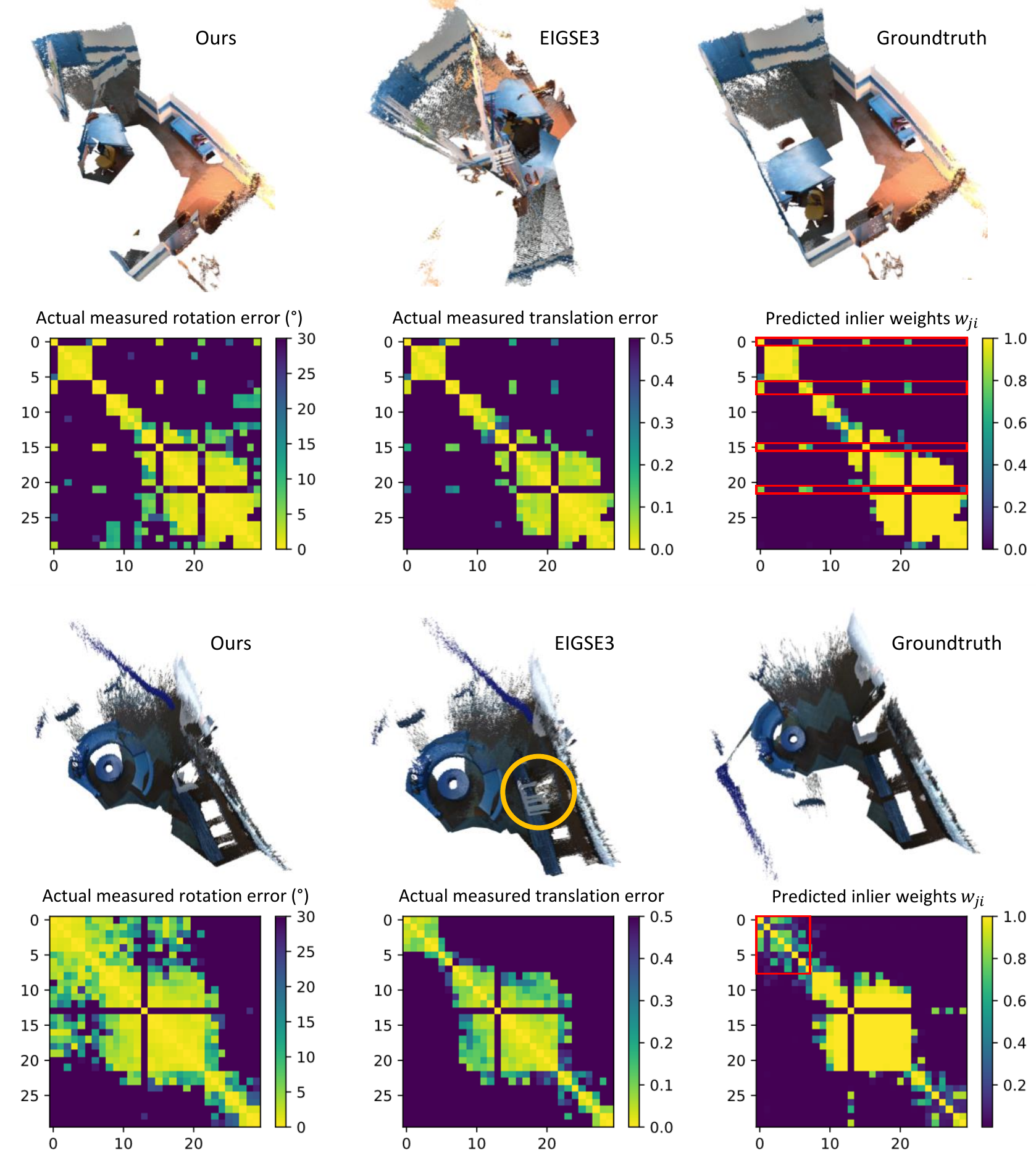}
    \caption{Example failure cases by our algorithm and EIGSE3 \cite{arrigoni2016eigse3}, which are caused by poor input pairwise transformations due to low overlap between frames. The resulting view-graphs are split into several disconnected components without accurate pairwise registrations between them, and we highlight one such separate component in red in the predicted adjacency matrices. Our algorithm still gives visually better results than EIGSE3.
    Top: scene0043\_00, bottom: scene0334\_02.}
    \label{fig:failures}
\end{figure*}

Fig. \ref{fig:success} shows additional qualitative results for multiway point cloud registration for ScanNet dataset using FCGF descriptors. The same examples are also shown in the accompanying video. We also show example failure cases in Fig. \ref{fig:failures}, which are due to small overlap between some point cloud pairs. In such cases, our algorithm still generates consistent alignments within each strongly connected component.

\end{document}